\documentclass[a4paper]{article}
\pdfoutput=1
\usepackage{style/INTERSPEECH2016}

\usepackage{graphicx}
\usepackage{subfigure}
\usepackage{amssymb,amsmath,bm}
\usepackage{textcomp}

\usepackage{tikz}
\usetikzlibrary{arrows,patterns}

\ninept

\title{Trainable Frontend For Robust and Far-Field Keyword Spotting}

\makeatletter
\def\name#1{\gdef\@name{#1\\}}
\makeatother \name{{\em Yuxuan Wang, Pascal Getreuer, Thad Hughes, Richard F. Lyon, Rif A. Saurous}}

\address{Google, Mountain View, USA \\
  {\small \tt \{yxwang,getreuer,thadh,dicklyon,rif\}@google.com}
}

\begin{document}

\maketitle

\begin{abstract}

Robust and far-field speech recognition is critical to enable true hands-free communication. In far-field conditions, signals are attenuated due to distance. To improve robustness to loudness variation, we introduce a novel frontend called per-channel energy normalization (PCEN). The key ingredient of PCEN is the use of an automatic gain control based dynamic compression to replace the widely used static (such as log or root) compression. We evaluate PCEN on the keyword spotting task. On our large rerecorded noisy and far-field eval sets, we show that PCEN significantly improves recognition performance. Furthermore, we model PCEN as neural network layers and optimize high-dimensional PCEN parameters jointly with the keyword spotting acoustic model. The trained PCEN frontend demonstrates significant further improvements without increasing model complexity or inference-time cost.
\end{abstract}

\section{Introduction}
Speech has become a prevailing interface to enable human-computer interaction, especially on mobile devices. An important component of such an interface is the keyword spotting (KWS) system~\cite{chen2014small}. For example, KWS is often used to wake up mobile devices or to initiate conversational assistants. Therefore, reliably recognizing keywords, regardless of the acoustic environment, is often a prerequisite for effective interaction with speech-enabled products.

Thanks to the development of deep neural networks (DNN), automatic speech recognition (ASR) has dramatically improved in the past few years~\cite{hinton2012deep}. However, while current ASR systems perform well in relatively clean conditions, their robustness remains a major challenge. This also applies to the KWS system~\cite{prabhavalkar2015automatic}. The system needs to be robust to not only various kinds of noise interference but also varying loudness (or sound level). The capability to handle loudness variation is important because it allows users to talk to devices from a wide range of distances, enabling true hands-free interaction.

Similar to modern ASR systems, our KWS system is also neural network based~\cite{chen2014small,sainath2015convolutional}. However, being a resource-limited embedded system, the keyword recognizer has its own constraints. Most importantly, it is expected to run on devices and is always listening, which demands small memory footprints and low power consumption. Therefore, the size of the neural network needs to be much smaller than those used in modern ASR systems~\cite{sainath2015convolutional,chen2015locally}, implying a network with a limited representation power. In addition, on-device keyword spotting typically does not use sophisticated decoding schemes or language models~\cite{chen2014small}. These constraints motivate us to rethink the design of the feature-extraction frontend.

In DNN-based acoustic modeling, perhaps the most widely used frontend is the so-called log-mel frontend, consisting of mel-filterbank energy extraction followed by log compression, where the log compression is used to reduce the dynamic range of filterbank energy. However, there are several issues with the log function. First, a log has a singularity at 0. Common methods to deal with the singularity are to use either a clipped log (i.e. $\log(\max(\text{offset}, x))$) or a stabilized log (i.e. $\log(x+\text{offset})$). However, the choice of the offset in both methods is ad hoc and may have different performance impacts on different signals. Second, the log function uses a lot of its dynamic range on low level, such as silence, which is likely the least informative part of the signal. Third, the log function is loudness dependent. With different loudness, the log function can produce different feature values even when the underlying signal content (e.g. keywords) is the same, which introduces another factor of variation into training and inference. Although techniques such as mean--variance normalization~\cite{chen2007mva} and cepstral mean normalization~\cite{atal1974effectiveness} can be used to alleviate this issue to some extent, it is nontrivial to deal with time-varying loudness in an online fashion.

To remedy the above issues of log compression, we introduce a new frontend called per-channel energy normalization (PCEN). Essentially, PCEN implements a simple feed-forward automatic gain control (AGC)~\cite{lyon2011cascades, perez2011automatic}, which dynamically stabilizes signal levels. Since all the PCEN operations are differentiable, we further propose to implement PCEN as neural network operations/layers and jointly optimize various PCEN parameters with the KWS acoustic model. Equipped with this trainable AGC-based frontend, the resulting KWS system is found to be more robust to distant speech.

The rest of the paper is organized as follows. In Section~\ref{pcen}, we introduce and describe the PCEN frontend. In Section~\ref{trainable_pcen}, we formulate PCEN as neural network layers and discuss the advantages of the new formulation. In Section~\ref{experiments}, we present experimental results. The last section discusses and concludes this paper.

\begin{figure}[t]
%\centering
\subfigure[log-mel frontend]{
\includegraphics[scale=0.34]{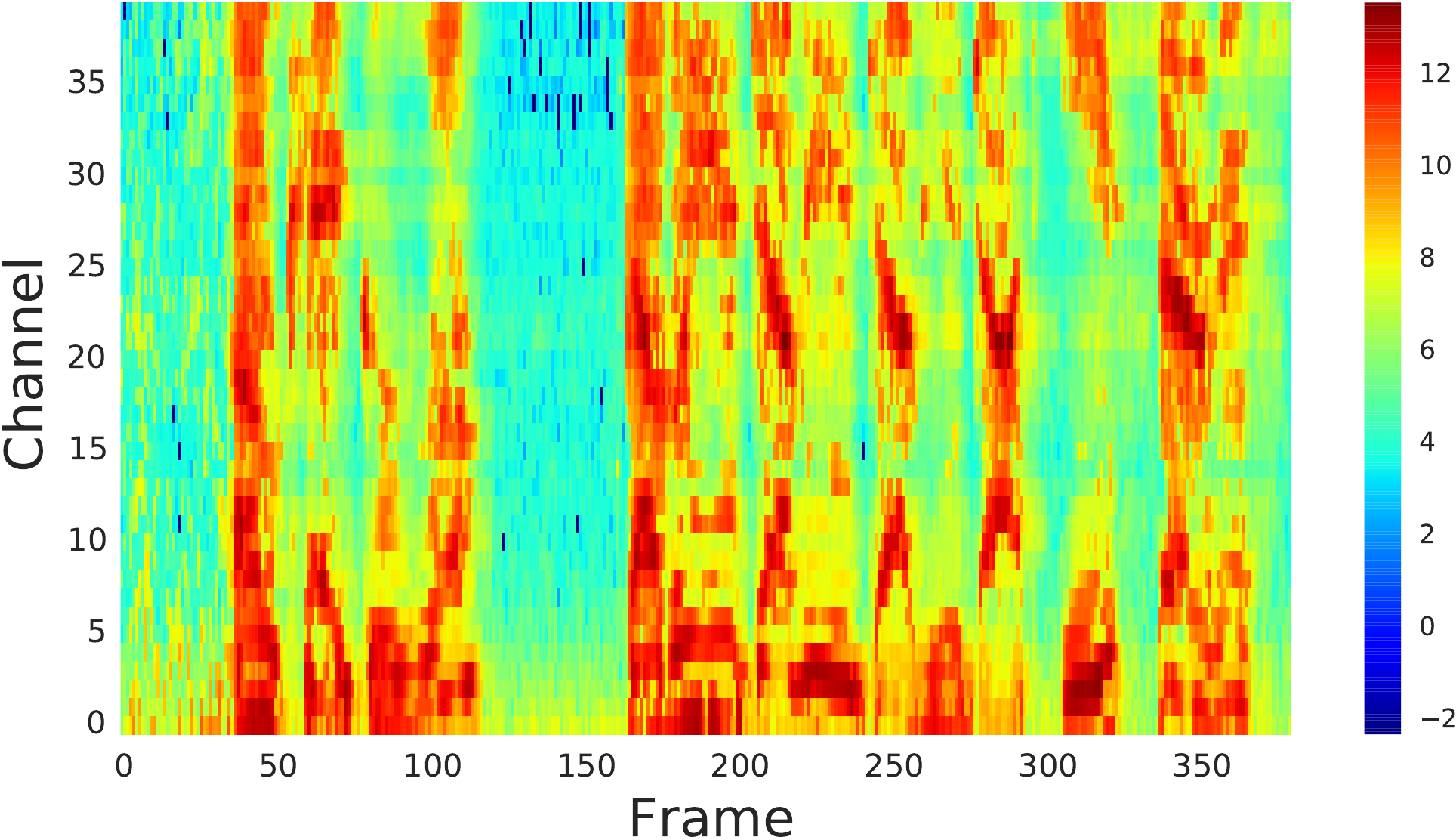}
}
%\hfil
\subfigure[PCEN frontend]{
\includegraphics[scale=0.34]{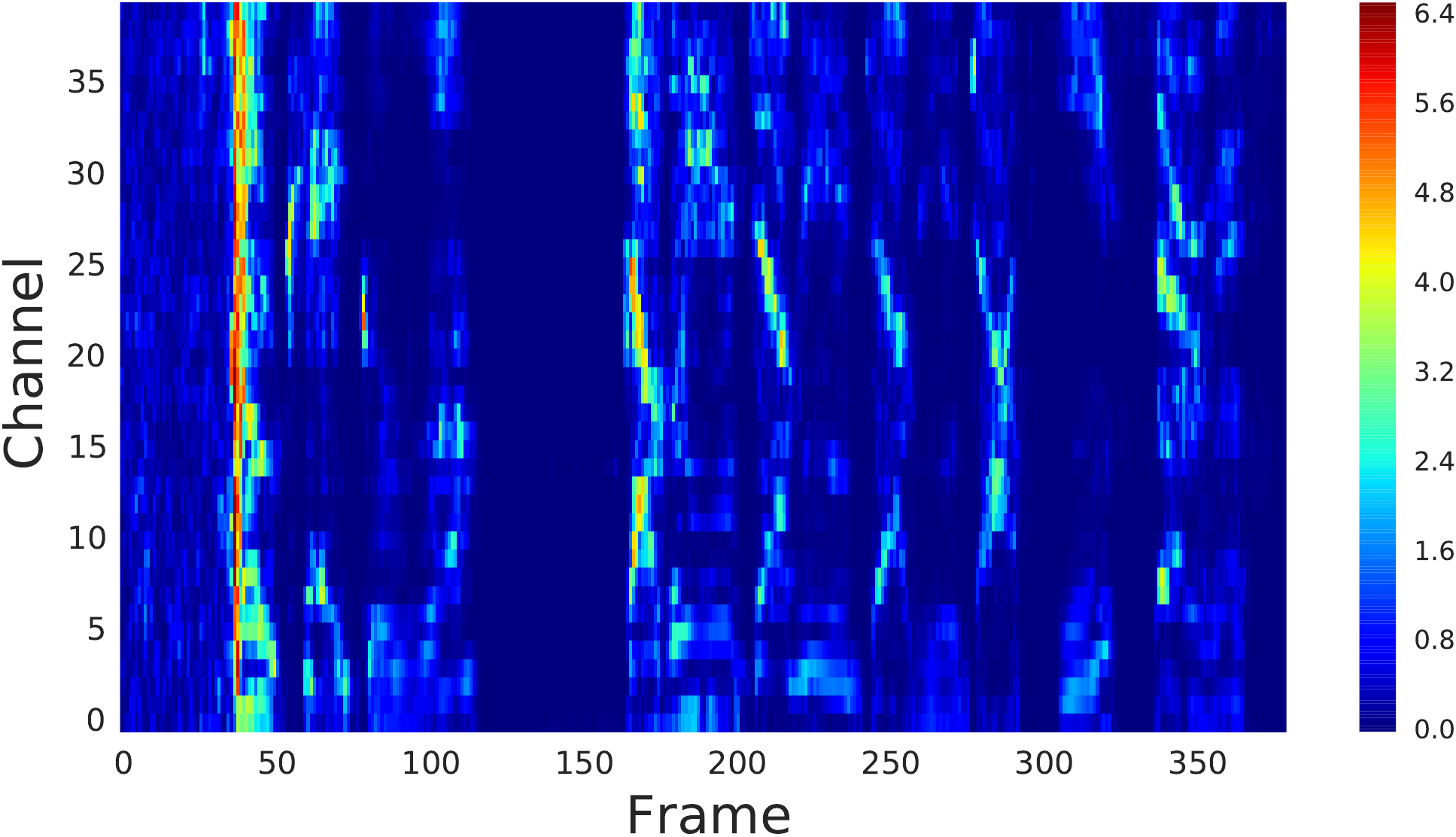}
}
\caption{{\it log-mel and PCEN features on a speech utterance.}}
\label{fig:log-mel_vs_pcen}
\end{figure}

\section{Per-Channel Energy Normalization}
\label{pcen}
In this section, we introduce the PCEN frontend as an alternative to the log-mel frontend. The key component in PCEN is that it replaces the static log (or root) compression by a dynamic compression described below:
\begin{equation}
{PCEN}(t, f) = \left(\frac{E(t, f)}{\left(\epsilon + M(t, f)\right)^{\alpha}} + {\delta}\right)^r - {\delta}^r,
\label{eq.pcen}
\end{equation}
where $t$ and $f$ denote time and frequency index and $E(t,f)$ denotes  filterbank energy in each time-frequency (T-F) bin. Although there is no restriction on what filterbank to use, in this paper we use an FFT-based mel filterbank for fair comparison with the log-mel frontend. $M(t,f)$ is a smoothed version of the filterbank energy $E(t,f)$ and is computed using a first-order infinite impulse response (IIR) filter, 
\begin{equation}
M(t, f) = (1-s) M(t-1, f) + s E(t, f),
\end{equation}
where $s$ is the smoothing coefficient. $\epsilon$ is a small constant to prevent division by zero. We arbitrarily chose $10^{-6}$ here and found it does not have significant performance impact. Essentially, the ${E(t, f)}/{(\epsilon + M(t, f))^{\alpha}}$ part implements a form of feed-forward AGC~\cite{perez2011automatic}. The AGC strength (or gain normalization strength) is controlled by the parameter $\alpha \in [0, 1]$, where larger $\alpha$ indicates stronger gain normalization. Note that due to smoothing, $M(t,f)$ mainly carries the loudness profile of $E(t,f)$, which is subsequently normalized out. Also note that this operation is causal and is done for each channel independently, making it suitable for real-time implementation. The AGC emphasizes changes relative to recent spectral history and adapts to channel effects including loudness~\cite{lyon2011cascades}. 

Following the AGC, we perform a stabilized root compression to further reduce the dynamic range using offset $\delta$ and exponent $r$. We note that the offset $\delta$ introduces a flat start to the stabilized root compression curve, which resembles an optimized spectral subtraction curve~\cite{porter1984optimal}. It is worth noting that the main parameters in PCEN are the AGC strength $\alpha$ and smoothing coefficient $s$, whose choices depend on the loudness distribution of data.

% Although it appears that PCEN introduces several parameters, it is worth noting that the main parameters in PCEN are the AGC strength $\alpha$ and smoothing coefficient $s$, whose choices depend on the loudness distribution of data. %After determining a good AGC strength, the offset $\delta$ and exponent $r$ are relatively easy to choose and there is a wide range of values that work well across different data sets. 

Figure~\ref{fig:log-mel_vs_pcen} compares the log-mel feature with PCEN feature on a speech utterance, where log-mel uses stabilized log with $\text{offset} = 0.1$ and PCEN uses $s = 0.025, \alpha = 0.98, \delta = 2$ and $r = 0.5$. Note that $s = 0.025$ translates to a 40-frame time constant for the IIR smoother. We can see that similar to log-mel, PCEN reduces the dynamic range of filterbank energy while still preserving prominent speech patterns. However, note that in several intervals (e.g. the first 30 frames and frame 120 to 170), low-level signals that do not carry useful information are amplified by the log operation, whereas in PCEN they are relatively flat. In addition, PCEN tends to enhance speech onsets, which are important for noise and reverberation robustness. One way to understand the onset enhancement effect is to rewrite ${E(t, f)}/{(\epsilon + M(t, f))^{\alpha}}$ as $\exp\bigl(\log(E(t,f)) - \alpha\log(\epsilon+M(t,f))\bigr)$. That is, PCEN can be interpreted as performing a partial log-domain high-pass filtering followed by an exponential expansion, which helps enhance transitions. This is similar to the RASTA filtering used for channel normalization~\cite{hermansky1994rasta}.

%\begin{figure}[t]
%\centering
%\includegraphics[width=0.95\linewidth]{plot/trainable_pcen.pdf}
%%\includegraphics[scale=0.45]{plot/trainable_pcen.pdf}
%\caption{{\it Schematic diagram of the trainable PCEN frontend. The acoustic model and frontend are jointly trained using backpropagation.}}
%\label{fig:trainable_pcen}
%\end{figure}

\begin{figure}[t]
\centering
\mbox{
%\scalebox{1.0}{
\beginpgfgraphicnamed{plots/trainable_pcen}
% TikZ figure: block diagram of trainable PCEN net architecture.

\begin{tikzpicture}[scale=0.62,yscale=-1.1,rounded corners=0.7mm,
>=stealth',shorten >=1pt, font=\sffamily\footnotesize]

\node [rectangle,draw,black] (A1) at (0,0) {KWS acoustic model};
\node [rectangle,draw,black] (B1) at (0,1.1)
{PCEN operations with trainable parameters};

\node [rectangle,draw,black,align=center] (C1) at (-3,2.5)
{Uncompressed \\ filterbank energy};
\node [rectangle,draw,black] (C2) at (2.6,2.5) {Learned smoother};

\node [rectangle,draw,black] (S1) at (-0.7,4) {Smoother 1};
\node [rectangle,draw,black] (S2) at (2.6,4) {Smoother 2};
\node [rectangle,draw,black] (S3) at (6.4,4) {Smoother $k$};

\begin{scope}[semithick]
\draw [->] (B1) to (A1);
\draw [->] (C1) to (B1);
\draw [->] (C2) to (B1);
\draw [->] (S1) -- node [left=8pt,midway] {$w_1(f)$} (C2);
\draw [->] (S2) -- node [left=-3pt,midway] {$w_2(f)$} (C2);
\draw [->] (S3) -- node [right=6pt,midway] {$w_k(f)$}
node [left=5pt,midway] {$\cdots$} (C2);
\end{scope}

\draw (4.5,4) node {$\cdots$};

\end{tikzpicture}
\endpgfgraphicnamed
%}
}
\caption{{\it Schematic diagram of the trainable PCEN frontend. The acoustic model and frontend are jointly trained using backpropagation.}}
\label{fig:trainable_pcen}
\end{figure}
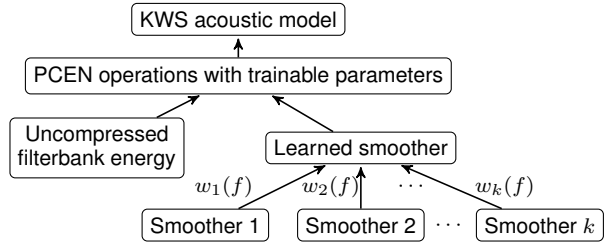

\section{Trainable PCEN Frontend}
\label{trainable_pcen}
As can be seen in Equation~\ref{eq.pcen}, PCEN introduces several parameters. These parameters need to be manually tuned, which is a labor intensive and inherently suboptimal process. In some cases, manual tuning can become impossible. For example, when one wants to jointly tune all parameters with a moderate resolution, the number of parameter combinations can easily explode. One of our original motivations is to find a way to automatically optimize the various PCEN parameters together, such that human efforts can be reduced.

Fortunately, all the PCEN operations are differentiable, permitting gradient-based optimization w.r.t.\ hyperparameters. We first describe how to learn gain normalization related parameters $\alpha, \delta$ and $r$. With precomputed filterbank energy $E$ and the corresponding smoother $M$, PCEN can be easily expressed as matrix operations, hence representable with standard neural network layers. Taking advantage of the modularity of the neural network, we can feed the PCEN layer outputs as features to the original KWS acoustic model and jointly optimize them. Given the KWS acoustic model loss function, we can use backpropagation to compute the gradient w.r.t.\ those parameters and update them iteratively using stochastic gradient descent (SGD). To ensure parameter positivity, we do gradient updates on their log values and then take exponentials.

The neural-network formulation offers more than automatic parameter finding. In the standard PCEN formulation (Equation~\ref{eq.pcen}), we restrict the parameters to be scalars shared across all T-F bins and manually tune them, because it is difficult to design high-dimensional parameters. With the neural-network-based formulation, we can generalize PCEN parameters to be frequency- or even time-frequency dependent and let SGD optimize their values automatically. 

We can extend this idea to learning the smoother $M$ as well. Specifically, we would like to learn the smoothing coefficient $s$, which controls the time constant of the IIR smoother. One way to achieve this would be to model the smoother as a recurrent neural network with a trainable $s$. Another way is to predetermine several smoothing coefficients and learn a weighted combination of several smoother outputs. In this work, we choose the second method and learn a frequency-dependent convex combination, i.e. (time index $t$ is omitted here)
\begin{align}
&M(f) = \sum_{k=1}^{K} w_k(f) M_k(f) \nonumber \\
&\text{s.t. } w_k(f) \ge 0 \text{ and } \sum_{k=1}^{K} w_k(f) = 1,
\label{eq.softmax}
\end{align}
where $K$ is the number of smoothers. The constraints in Equation~\ref{eq.softmax} can be satisfied by computing $w_k(f)$ as  $\exp\left(z_k(f)\right) / \sum_k\exp\left(z_k(f)\right)$ where $z_k(f) \in \mathbb{R}$. Similar to above, we can compute gradient w.r.t.\ all $z_k(f)$ and update them iteratively. 

Figure~\ref{fig:trainable_pcen} illustrates the joint PCEN frontend and KWS model. Here, we would like to point out an important design choice. In our design, all the trainable parameters are data-independent, meaning that they are not conditioned on any input features (that is, they are the ``leaf" nodes in the computation graph). After training, the learned parameters are frozen and do not change for different signals. This design choice avoids introducing additional network weights and computational cost during inference, because we can simply hardcode the trained parameters into the PCEN frontend. Although it is entirely possible that data-dependent parameter learning can further improve performance, it is important to strike a balance between accuracy and resource usage for always-listening KWS.

\section{Experiments}
\label{experiments}
\subsection{Experimental setup}
Our KWS system uses a convolutional neural network (CNN) \cite{sainath2015convolutional}. The input to the CNN consists of 23 left context frames and 8 right context frames (32 frames in total). For both log-mel and PCEN frontend, each frame is computed based on a 40-channel mel filterbank with 25ms windowing and 10ms frame shift. The CNN consists of a convolutional layer with 308 feature maps and $8\times8$ non-overlapping kernels, followed by a linear projection layer of size 32 and a fully connected rectified linear unit layer of size 128. The CNN is trained on large internal training sets to detect the keyword ``Ok Google". The interested reader is referred to~\cite{chen2014small, prabhavalkar2015automatic, sainath2015convolutional} for more details such as decoding.

To improve noise robustness, we perform multi-condition training by artificially corrupting each utterance with various interfering background noises and reverberations, where the noise sources contain sounds sampled from daily-life environments and YouTube videos. To further improve loudness robustness, we also perform multi-loudness training by scaling the loudness of each training utterance to a randomly selected level ranging from $-45$~dBFS to $-15$~dBFS. The far-field eval sets used in this paper are rerecorded in real environments (such as a moving car or a living room). The negative data in these sets mostly contain speech signals mined from real voice search logs. The size of the eval sets ranges from about 50 to 300 hours. There is no overlapping condition between the training and eval sets. All the data used are anonymized.

Our evaluation metric uses the receiver operating characteristic (ROC) curve, plotting false rejection (FR) rates against false alarm (FA) rates. Our goal is to achieve low FR rates while maintaining extremely low FA rates (e.g. no more than 0.5 false alarms per hour of audio).

\subsection{PCEN vs. log-mel}
\begin{figure}[t]
\centering
\includegraphics[width=0.8\linewidth]{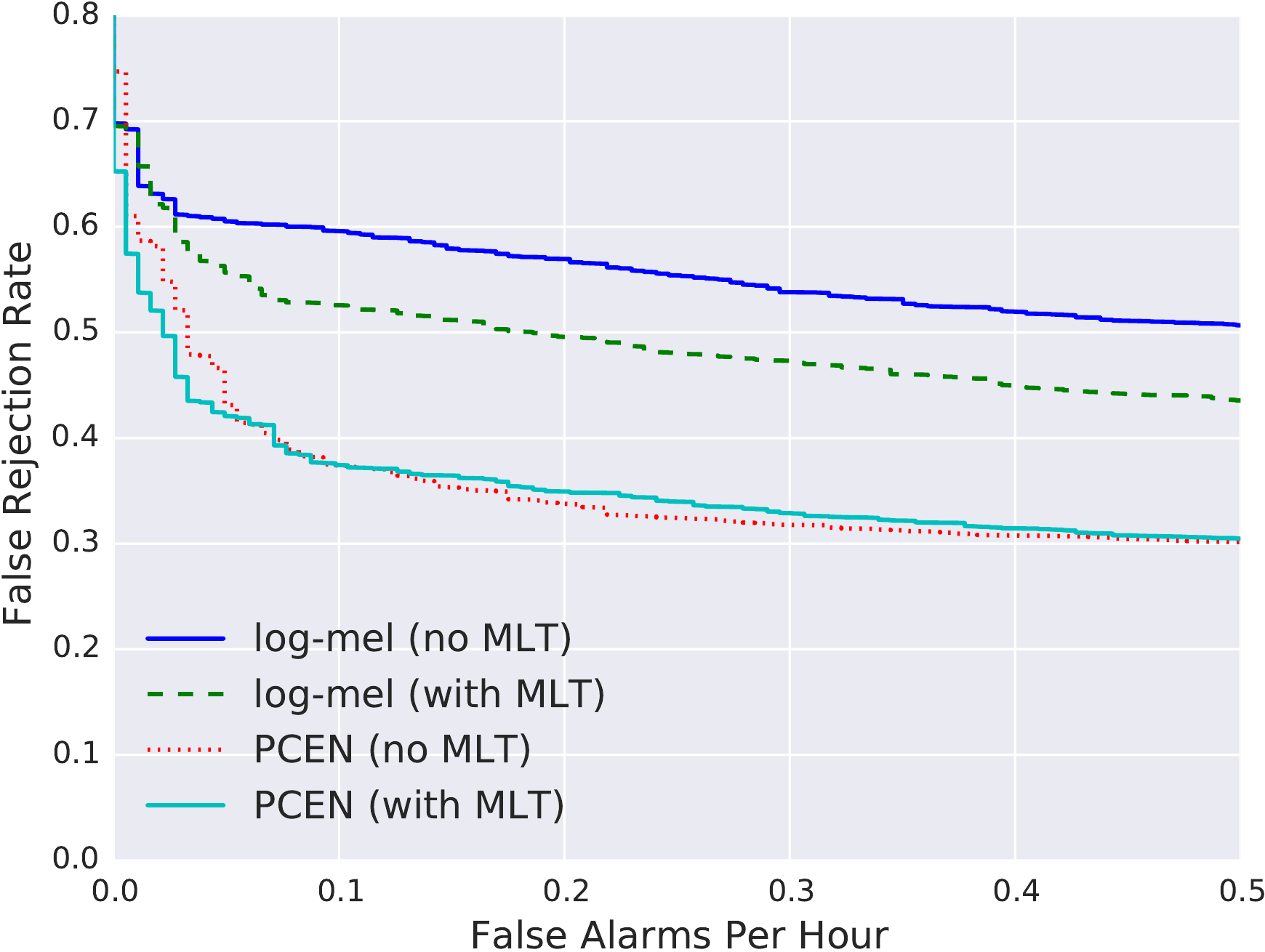}
\caption{{\it ROC curve comparison between log-mel and PCEN on a difficult far-field eval set. ``MLT" stands for multi-loudness training.}}
\label{fig:car_rerecorded}
\end{figure}

We first present results comparing the log-mel frontend with the proposed PCEN frontend. Note that we compare a log-mel frontend containing an optimized spectral subtraction process (details omitted here), which already outperforms the standard log-mel frontend. In this experiment, PCEN uses a fixed set of scalar parameters, where $s=0.025, \alpha=0.98, \delta=2$ and $r=0.5$. To better demonstrate the gain normalization effect of PCEN, we compare both frontends with and without multi-loudness training. Fig.~\ref{fig:car_rerecorded} shows the ROC comparison on a rerecorded far-field car eval set. We can see that log-mel without multi-loudness training performs the worst, and adding loudness variations into training clearly helps log-mel. In contrast, the proposed PCEN frontend significantly outperforms log-mel even when it is not multi-loudness trained. For example, at 0.1 FA per hour, PCEN reduces FR rate by about 14\% absolute over log-mel with multi-loudness training. Interestingly, PCEN with and without multi-loudness training perform similarly to each other on this eval set, which is likely due to the use of a high AGC strength $\alpha=0.98$. Nevertheless, we find that multi-loudness training helps in general and is needed for the trainable version of PCEN, hence the rest of the experiments all use multi-loudness training.

We emphasize that multi-condition training on large amounts of data is a very strong baseline, hence the obtained improvements are quite significant. In addition, we have also compared with RASTA filtering~\cite{hermansky1994rasta}, a widely used channel normalization technique, but found its performance to be inferior (not shown here).

\begin{figure}[th!]
\centering
\subfigure[Noisy and far-field eval set (0.8m distance)]{
\includegraphics[scale=0.335]{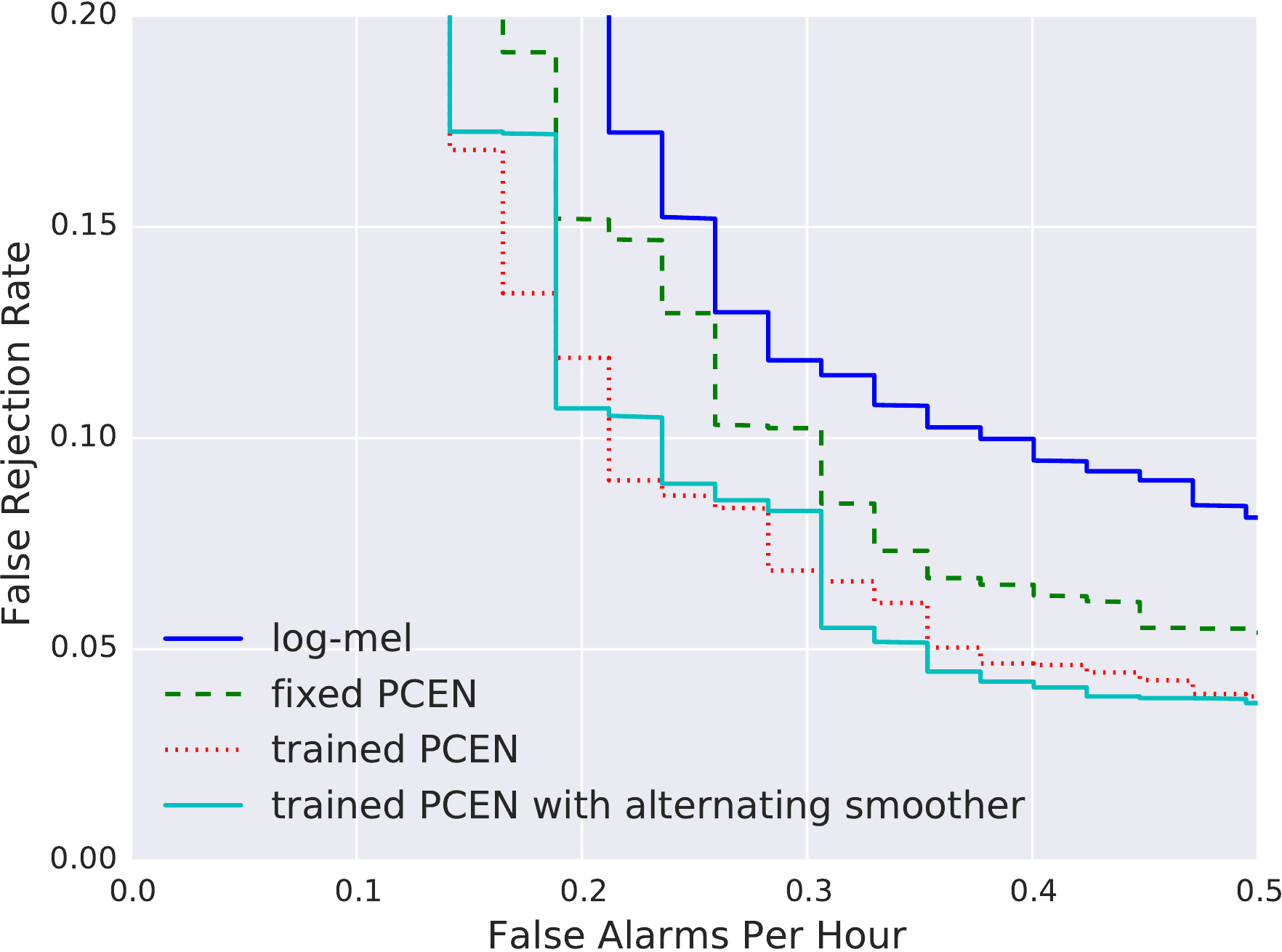}
\label{fig:eval_80cm}
}
\subfigure[Noisy and far-field eval set (5m distance)]{
\includegraphics[scale=0.335]{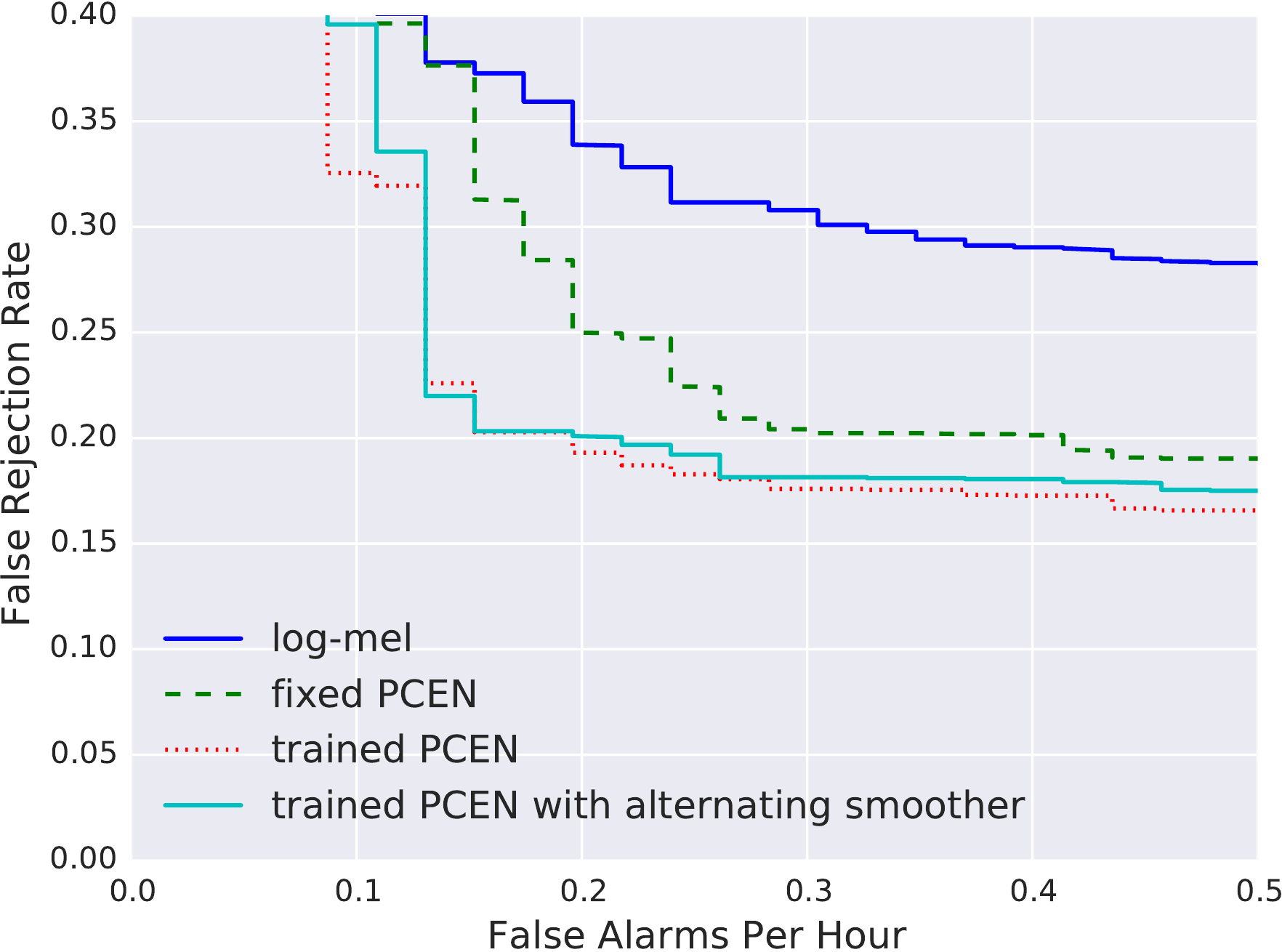}
\label{fig:eval_500cm}
}
\subfigure[Clean and near-field eval set]{
\includegraphics[scale=0.335]{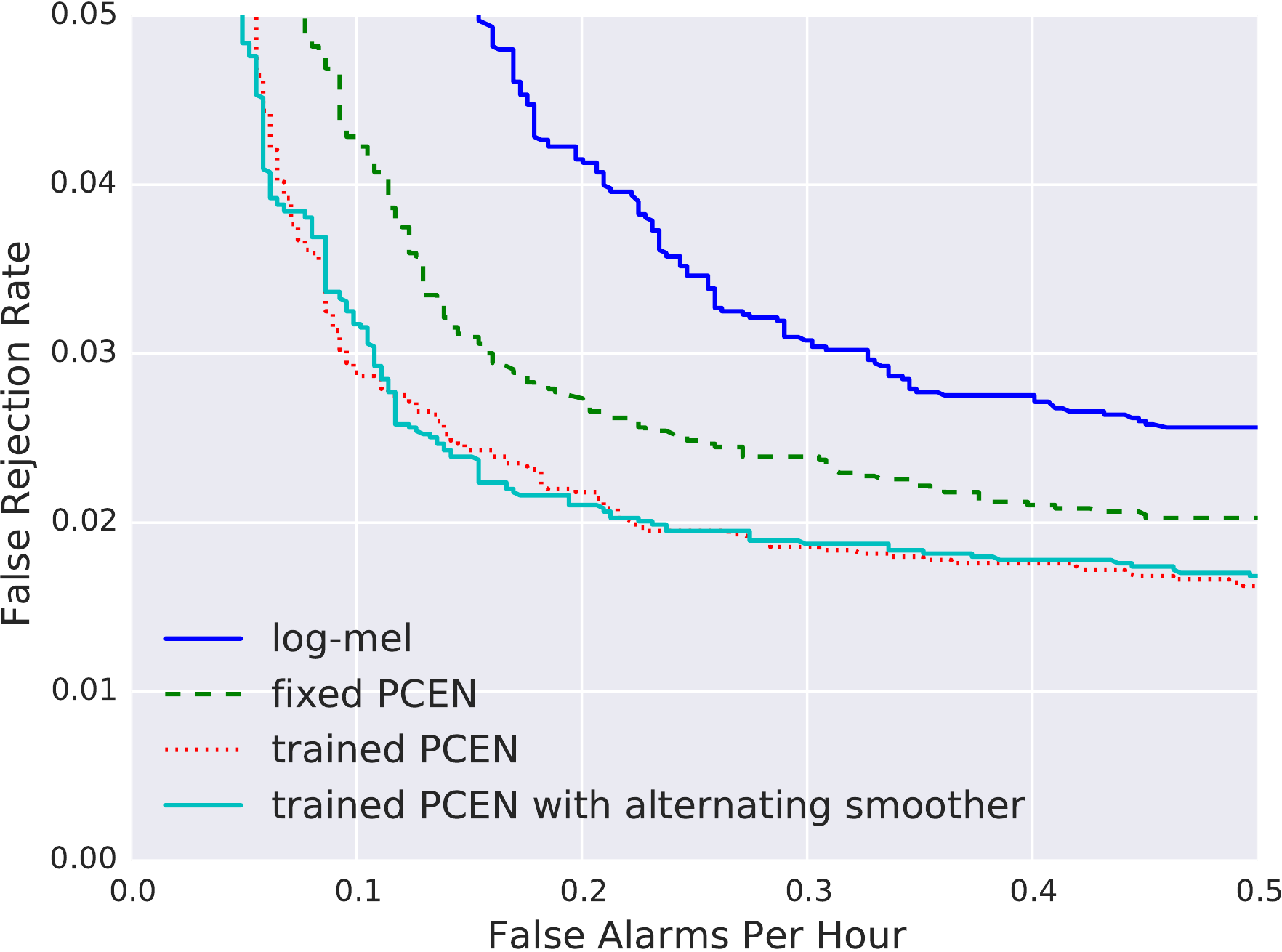}
\label{fig:eval_clean}
}
\caption{ROC curves of log-mel, fixed PCEN and trained PCEN. All models are trained with multi-loudness data.}
\label{fig:farfield_eval}
\end{figure}

\subsection{Trainable PCEN}
In this subsection, we compare fixed and trained PCEN. As mentioned above, in trainable PCEN, we generalize parameters to be frequency-dependent and jointly optimize all of them, including $\alpha(f), \delta(f), r(f)$ and $z_k(f)$. Here, $\alpha(f), \delta(f)$ and $r(f)$ are randomly initialized from a normal distribution with mean $1.0$ and standard deviation $0.1$. $z_k(f)$ is randomly initialized from a normal distribution with mean $\log(1/K)$ and standard deviation $0.1$, where $K$ is the number of predetermined smoothers. In our preliminary experiments, we computed 4 smoothers using smoothing coefficients of $0.015$, $0.02$, $0.04$, and $0.08$. We observed that the learned weights are mostly assigned to the slowest and fastest smoothers (that is, $s=0.015$ and $s=0.08$). Therefore, we use only those two smoothers in the following experiments.

We note two training details. First, for effective training of PCEN, we found that it is important to have multi-loudness training data, such that the network can learn the importance of gain normalization. Furthermore, we found that it is helpful to bootstrap the KWS CNN layers from a pretrained PCEN model.

Figure~\ref{fig:eval_80cm} and \ref{fig:eval_500cm} show performance comparisons on two rerecorded far-field eval sets, where the talking distances are 0.8m and 5m, respectively. We can see that the trained PCEN consistently outperforms fixed PCEN on both eval sets. Compared to the multi-loudness trained log-mel, the improvements from trained PCEN are quite significant, especially in low FA regions. An interesting question is whether the proposed frontend would actually hurt the performance in clean and near-field conditions. As can be seen in Fig.~\ref{fig:eval_clean}, both fixed PCEN and trained PCEN outperform log-mel on a large clean and near-field eval set.

\subsection{Learned smoother combination weights}

\begin{figure}[!]
\centering
\includegraphics[scale=0.45]{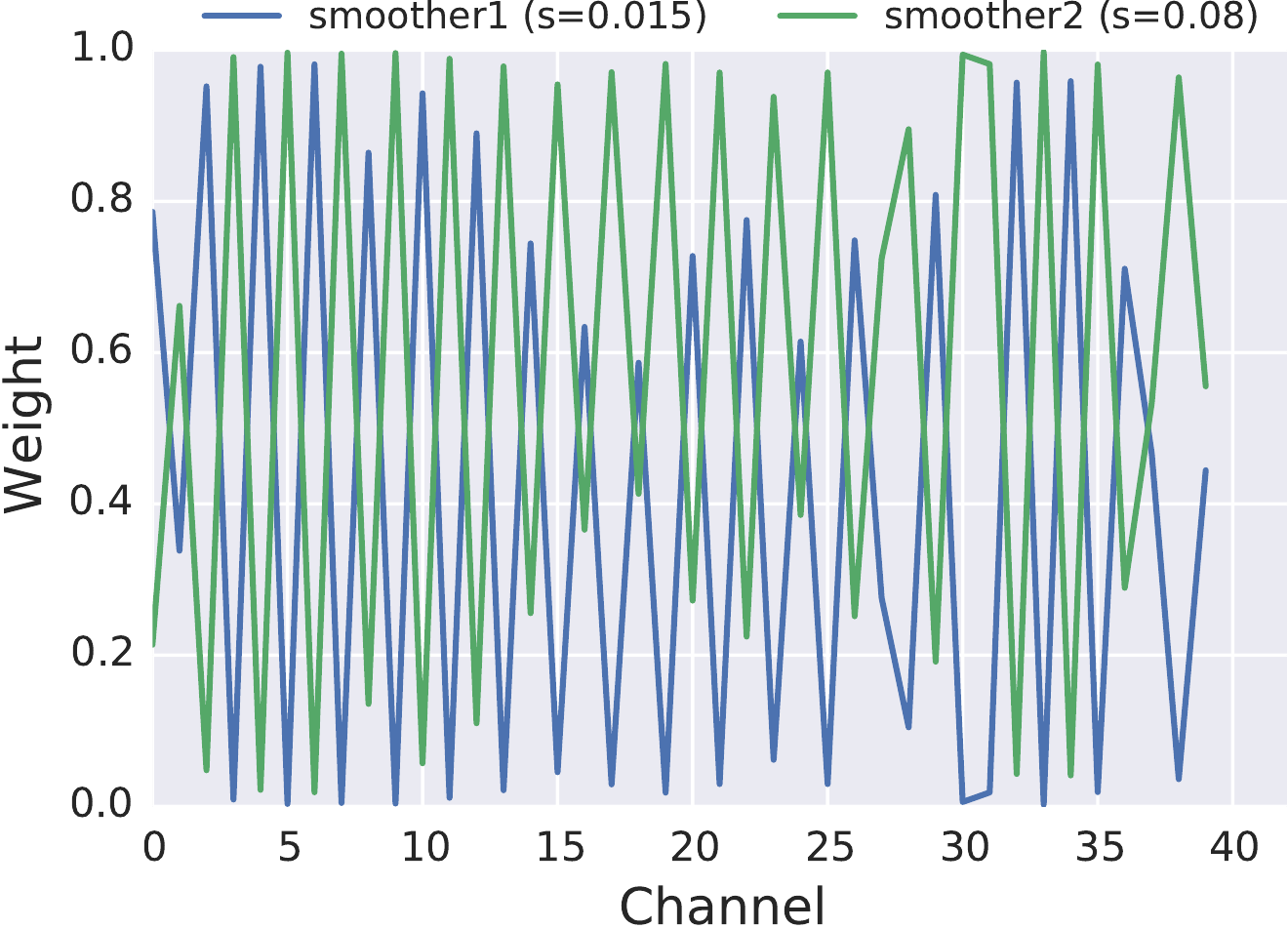}
\caption{{\it Learned per-channel weights of a slow ($s=0.015$) and fast ($s=0.08$) smoother.}}
\label{fig:alternate}
\end{figure}

Figure \ref{fig:alternate} shows an example of the smoother combination weights learned by the joint model. Interestingly, the learned weights show an alternating pattern between frequency channels, where even-numbered channels prefer to use the slower smoother ($s=0.015$) while odd-numbered channels the faster smoother ($s=0.08$). We suspect that this pattern is likely caused by the redundancy in filterbank energy from neighboring channels. The network basically tries to alternate between different features to obtain more discriminative information. This pattern could also be tied to the specific CNN architecture used. In the future, we would like to design experiments to better understand this phenomenon.

The architecture shown in Fig.~\ref{fig:trainable_pcen} increases computational cost because it needs to compute multiple smoothers first. In fact, the alternating pattern can be utilized to reduce the cost. Inspired by the spiky alternating weights, we trained a new model with a single smoother but computed using alternating smoothing coefficients (the other parameters are still learned as described before). As shown in Fig.~\ref{fig:farfield_eval}, the performance of the resulting model is actually very similar to the full trained one. This result has an important implication because it shows that we can achieve better performance without increasing inference-time complexity.

\section{Discussions and Conclusions}
\label{conclusion}
We have introduced a new robust frontend called PCEN, where the key idea is to replace the static log (or root) compression by an AGC-based dynamic compression. PCEN is conceptually simple, computationally cheap, and easy to implement. It significantly outperforms the widely used log-mel frontend in noisy and far-field conditions. 

We have also formulated PCEN as neural network layers. This formulation not only allows us to perform end-to-end training but also to generalize PCEN to have frequency- or time-frequency dependent parameters. The resulting model provides significant further improvements without increasing inference-time complexity. 

To conclude, this work represents an effort to embed signal processing components into a general-purpose neural network model, where the signal processing components can be viewed as structural regularizations. We think this is a promising future research direction.
\\

\noindent\textbf{Acknowledgements:} The authors would like to thank Malcolm Slaney and Alex Gruenstein for helpful comments.

\eightpt

\bibliographystyle{style/IEEEtran}
\bibliography{pcen_bib}

\end{document}